%% file: main.tex
\documentclass[runningheads]{llncs}

\usepackage{eccv}

\usepackage{eccvabbrv}
\usepackage{multirow}
\usepackage{graphicx}
\usepackage{booktabs}

\usepackage[accsupp]{axessibility}  

\usepackage{hyperref}

\usepackage{orcidlink}

\begin{document}

\title{DSAR: Dual-Stream Autoregressive Modeling of Temporal Cloth Dynamics for Photorealistic Animatable Avatars}

\titlerunning{DSAR}


\author{Haozhong Xiong\inst{1}\orcidlink{0009-0002-4619-9804} \and
Yao Yu\inst{1}\orcidlink{0000-0002-1111-7939}\thanks{Corresponding authors.} \and
Yu Zhou\inst{1}\orcidlink{0000-0001-5345-8902}$^*$ \and
Sidan Du\inst{1}\orcidlink{0000-0002-9966-3765}}

\authorrunning{H.~Xiong et al.}


\institute{Nanjing University, Nanjing, China\\
\email{502024230011@smail.nju.edu.cn, \ \{allanyu,nackzhou,coff128\}@nju.edu.cn}}

\maketitle
\begin{center}
\includegraphics[width=\textwidth]{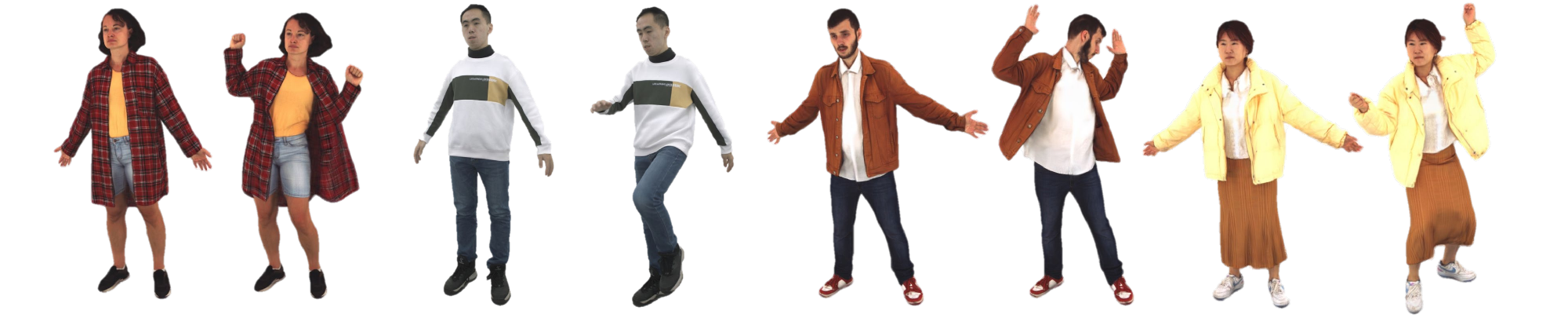}
\captionof{figure}{We propose DSAR, which explicitly models temporal causality in cloth dynamics through dual-stream autoregressive architecture, achieving realistic deformations on loose clothing and robust generalization to out-of-distribution poses.}
\label{fig:teaser}
\end{center}
\input{sec/0_abstract}    
\input{sec/1_intro}
\input{sec/2_related_works}
\input{sec/3_Method}

\input{sec/4_Experiment}

\input{sec/5_Conclusion}

%
%
\bibliographystyle{splncs04}
\bibliography{main}

\clearpage
\appendix
\input{Supplementary_Material}

\end{document}

%% file: sec/0_abstract.tex
\begin{abstract}
Creating photorealistic and temporally coherent animatable human avatars from RGB videos remains challenging. Current methods struggle to capture realistic cloth dynamics, producing over-smoothed appearance or severe artifacts on out-of-distribution poses. This limitation stems from a fundamental oversight: existing approaches neglect the temporal causality inherent in cloth physics, where current states emerge from previous states through temporal evolution rather than instantaneous skeletal configurations alone. Without explicit modeling of this causal structure, networks learn pose-appearance correlations instead of motion evolution, leading to poor generalization.
We introduce a dual-stream autoregressive framework that explicitly models both observable geometric information and implicit internal state. The geometric stream propagates surface displacement from the previous frame, while the state stream fuses current features with historical states retrieved from a memory bank. Motion-adaptive aggregation handles spatially-varying dynamics, and adaptive regularization balances smoothness with flexibility. Experiments on challenging datasets demonstrate significant improvements in rendering quality, temporal consistency, and generalization to motion patterns beyond training distributions, validating that dual-stream temporal modeling enables realistic cloth dynamics.

\keywords{Animatable avatars \and Neural rendering \and Dual-Stream}
\end{abstract}

%% file: sec/1_intro.tex
\section{Introduction}
\label{sec:intro}

The creation of photorealistic, animatable human avatars has emerged as a cornerstone technology for virtual reality, telepresence, film production, and interactive media. The ultimate goal is to capture the dynamic interplay between human motion and garment appearance—subtle wrinkles forming during arm bending, fabric momentum during rapid spinning, and gradual cloth settling after motion ceases. These temporal phenomena are fundamental to visual realism yet remain challenging for current neural rendering approaches.

Recent advances in neural rendering have enabled impressive progress in avatar creation from multi-view RGB videos. Building on foundational techniques in neural radiance fields~\cite{nerf,mipnerf} and 3D Gaussian Splatting~\cite{3dgs}, numerous methods~\cite{tava,arah,gaussianavatar,hugs,animatablegs} have achieved high-fidelity avatar reconstruction. Despite differences in underlying representations, these methods share a common modeling paradigm: they model garment appearance as functions of current or recent skeletal poses, where pose information is derived from either single frames or concatenated across temporal window. While this design enables efficient per-frame reconstruction, it exhibits fundamental limitations in practice—over-smoothed appearance on training poses, temporal flickering during motion, and severe artifacts including unrealistic wrinkles on novel pose sequences.

We identify that this limitation stems from ignoring the temporal causal structure inherent in cloth motion. In reality, garment appearance at time $t$ is not uniquely determined by skeletal pose alone. Consider a standing pose reached after rapid spinning versus from rest: while skeletal configurations are nearly identical, cloth states differ drastically due to motion history—the post-spinning garment exhibits residual momentum and dynamic wrinkles absent in the static case. This reveals a fundamental one-to-many mapping where identical poses correspond to different cloth states depending on temporal context.

Analysis of real-world cloth motion reveals that resolving this ambiguity requires two complementary types of information. The first is \textbf{explicit kinematic information}—the observable geometric configuration (where cloth surfaces are positioned) and motion patterns (how surfaces are moving, characterized by velocity and directional momentum), which together describe the current observable state of cloth surfaces. The second is \textbf{implicit internal state}, encompassing latent properties such as fabric tension, deformation history, and material memory that influence cloth appearance but cannot be directly observed from surface geometry alone. Only when both are accounted for can the cloth state at time $t$ be uniquely determined.

Skeletal poses provide only body joint positions, lacking both cloth kinematic information (geometric configuration and motion patterns) and internal state. This dual deficiency creates fundamental prediction ambiguity—without explicit cloth state feedback, networks must infer both "where and how cloth surfaces are moving" and "what internal factors govern their evolution" from skeletal motion alone, leading to the observed one-to-many mapping problem.

Various approaches have attempted to address this issue. RealityAvatar~\cite{realityavatar} and MonoHuman~\cite{monohuman} optimize per-frame latent codes to disambiguate identical poses during training, but rely on pose-based interpolation at test time that cannot capture motion-dependent variations. InstantAvatar~\cite{instantavatar} and HumanRF~\cite{humanrf} incorporate historical skeletal poses through concatenation or transformer aggregation to provide temporal context, but skeletal representations capture only body-centric motion—where the skeleton was but not how cloth has evolved—creating ambiguity when identical poses correspond to different cloth states. Test-time alignment strategies~\cite{animatablegs,ramavatar} improve generalization by mapping novel poses back to training distributions, but remain effective primarily when test poses exhibit small distributional shifts from training data.

Despite their diversity in architectural design, these learning-based avatar rendering methods do not establish explicit temporal causality between successive cloth states, instead learning correlations between pose patterns and appearance rather than temporal evolution rules.

Physics-based cloth simulation~\cite{snug,hood,cloth3d,diffcloth,neuralclothsimulation} naturally incorporates temporal causality through explicit state evolution, tracking cloth configurations across timesteps to model momentum and dynamic deformations. However, these methods are fundamentally limited by human modeling capabilities—the precision of manually-crafted material models, the completeness of physical constraints, and the approximations necessary for computational tractability may fail to capture the full complexity of real-world cloth behavior. Our method addresses cloth dynamics from a complementary data-driven perspective, learning temporal evolution rules directly from multi-view photometric observations. This image-based supervision enables the network to discover dynamics from visual evidence of real-world phenomena, potentially capturing effects that extend beyond manually-modeled physics.

We address this through explicit dual-stream modeling. The geometric stream establishes cloth geometric information propagation via autoregressive conditioning on the previous frame's predicted deformation $\Delta\mu_{t-1}$, providing explicit feedback on cloth surface configuration and movement rather than inferring them from body pose alone. The state stream maintains a memory bank of historical temporal states that encode implicit internal state, capturing latent properties that are difficult to model explicitly but govern temporal evolution. These two streams address the dual requirements: the geometric stream tracks observable kinematics deformation, while the state stream captures implicit internal state learned from visual data.

\noindent\textbf{Contributions.} Our work makes the following contributions:
\begin{itemize}
    \item We identify that temporally coherent cloth rendering requires modeling both observable kinematic information and implicit internal state, which cannot be inferred from skeletal poses alone due to the one-to-many mapping problem.
    
    \item We propose a dual-stream autoregressive architecture that explicitly addresses these dual requirements from photometric supervision: a geometric stream propagates kinematic information through autoregressive conditioning on previous frame's cloth deformation and motion-adaptive temporal aggregation, while a state stream encodes implicit internal state through memory-based retrieval of historical temporal features.
    
    \item Our framework achieves significant improvements in rendering quality, temporal consistency, and generalization to motion patterns beyond training distributions.
\end{itemize}

\begin{figure*}[t]
\centering
\includegraphics[width=\textwidth]{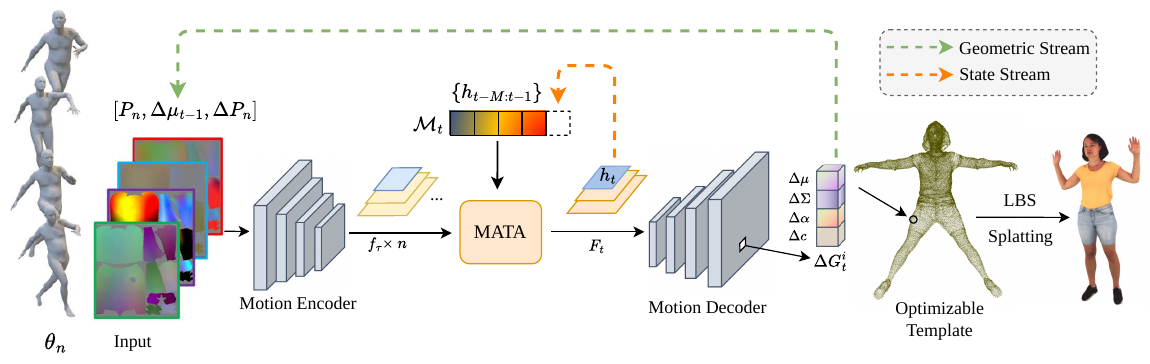}
\caption{\textbf{Overview.} Our dual-stream framework processes a temporal window and previous deformation $\Delta\mu_{t-1}$ through two complementary pathways. The geometric stream aggregates multi-scale features via MATA. The state stream fuses aggregated features with historical states from memory bank $\mathcal{M}_t$ to obtain temporal state $h_t$, which is decoded into Gaussian deformations $\Delta G_t$ and stored for next-frame inference.}
\label{fig:overview}

\end{figure*}

\begin{figure}[t]
\centering
\includegraphics[width=0.5\columnwidth]{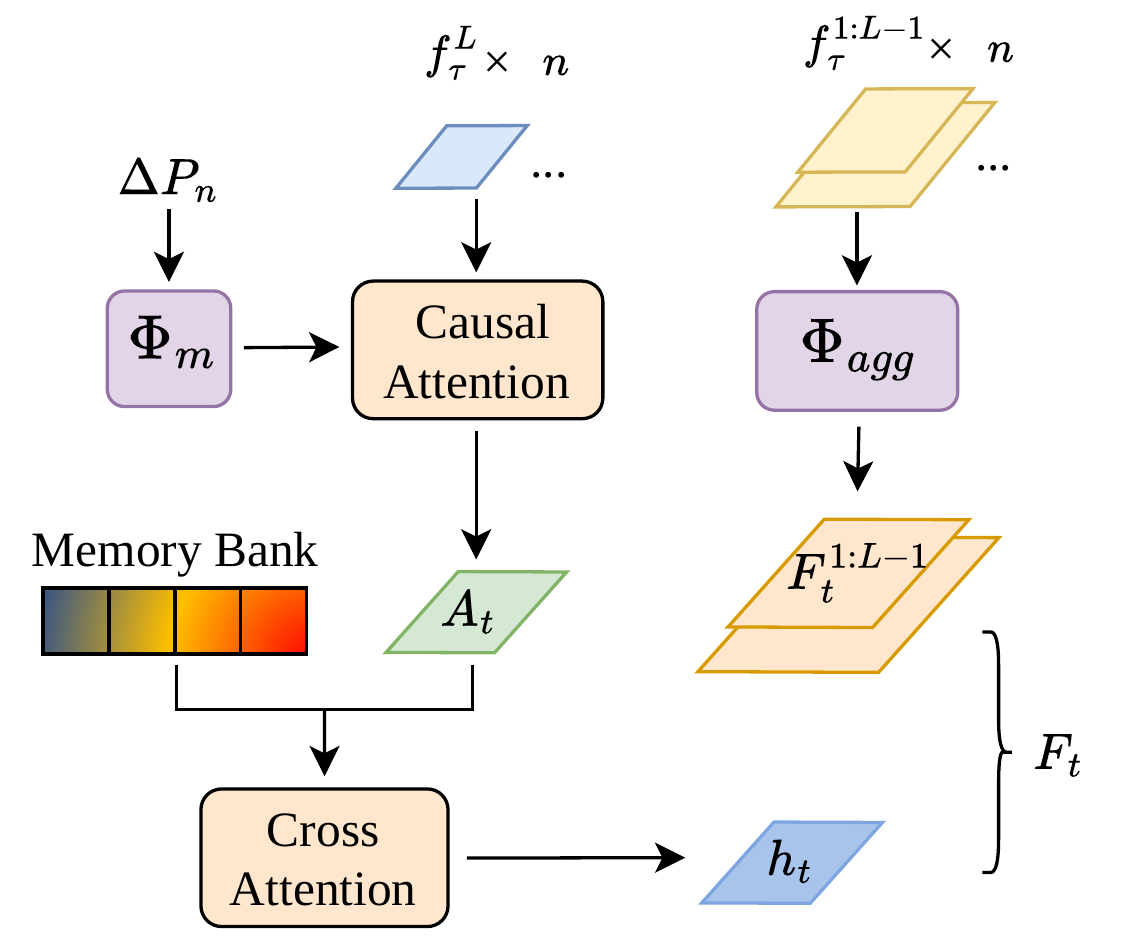}
\caption{\textbf{Motion-Adaptive Temporal Aggregation (MATA).} Features are aggregated hierarchically: motion-adaptive causal attention at the deepest level ($f_{\tau}^L$) and lightweight convolutions ($\Phi_{{agg}}$) at finer levels ($f_{\tau}^{1:L-1}$) .}
\label{fig:MATA}
\end{figure}

%% file: sec/2_related_works.tex
\section{Related Work}

\noindent\textbf{Animatable Human Avatar Representation.} Early reconstruction methods employ implicit functions, such as occupancy fields~\cite{pifu,pifuhd,arch,arch++,leap,coap,occupancy} and signed distance functions~\cite{pamir,pina,scanimate,neuralgif,metaavatar,deepsdf}, or utilize explicit mesh representations~\cite{DDC,deliffas,icon,econ} to reconstruct clothed humans from scans or depth sequences. However, these representations often struggle to model view-dependent appearance effects and complex materials, limiting photorealistic rendering quality and temporal consistency.

Building on neural rendering techniques~\cite{lombardi2019neural,sitzmann2019scene,sitzmann2020implicit,stateoftheart,dnerf}, neural radiance field-based methods have revolutionized avatar creation through differentiable volumetric rendering with view-dependent appearance. Recent advances~\cite{muller2022instant,chen2022tensorf,fridovich2022plenoxels} have enabled numerous human modeling approaches~\cite{tava,posevocab,neuralbody,arah,avatarrex,guo2023vid2avatar,su2021anerf,neuralactor}. HumanNeRF~\cite{humannerf} learns pose-dependent deformations using skeletal motion priors, while Neural Body~\cite{neuralbody} associates latent codes with SMPL vertices for pose-conditioned rendering. AnimatableNeRF~\cite{animatablenerf} establishes a canonical NeRF with pose refinement for pose-driven animation. TAVA~\cite{tava} adopts explicit warping fields for fine-grained control. However, these implicit representations are computationally expensive, and their MLP-based architectures struggle to capture high-frequency geometric details due to spectral bias~\cite{mlp}.

3D Gaussian Splatting~\cite{3dgs} enables real-time rendering through explicit point-based primitives, inspiring various avatar modeling approaches~\cite{gaussianavatar,3dgssurvey,splattingavatar,humangaussiansplatting,drivable,gauhuman}. GaussianAvatar~\cite{gaussianavatar} binds Gaussians to UV maps with learnable features for pose-driven animation. Animatable-GS~\cite{animatablegs} employs StyleUNet~\cite{StyleGAN,StyleUnet} for multi-scale deformation prediction with learnable skinning weights. To enhance geometric quality, several methods~\cite{gpsgaussian,vga} introduce geometric priors including as-rigid-as-possible constraints and normal consistency regularization. Despite achieving impressive rendering performance, these methods fundamentally model clothing appearance as functions of current or recent skeletal poses, formulated as $\text{Appearance}_t = F(\text{pose}_{t-n:t})$, neglecting the temporal causality where current cloth states emerge from previous states through evolution.

\noindent\textbf{Temporal Modeling in Neural Avatar Rendering.} Various neural rendering approaches have explored temporal modeling to capture cloth dynamics beyond instantaneous pose-to-appearance mappings.

Per-frame latent encoding approaches~\cite{realityavatar,monohuman,uvvolumes,structured} optimize frame-specific codes to disambiguate identical poses during training, capturing variability that skeletal pose alone cannot explain. RealityAvatar~\cite{realityavatar} learns per-frame latent codes jointly with neural rendering, while MonoHuman~\cite{monohuman} employs 4D human representation with spatio-temporal features. However, these codes are optimized independently without temporal constraints between frames. At test time, codes for novel poses must be inferred through pose-based interpolation—finding training frames with similar skeletal configurations and blending their optimized codes—which cannot capture motion-dependent variations where identical poses correspond to different cloth states depending on motion history.

Skeletal pose concatenation methods~\cite{instantavatar,humannerf,humanrf} incorporate historical skeletal poses as network input to provide temporal context through body motion sequences. InstantAvatar~\cite{instantavatar} concatenates multiple consecutive pose parameters through temporal encoders, while HumanRF~\cite{humanrf} employs transformers to aggregate pose sequences for dynamic human modeling. Neural Cloth Simulation~\cite{neuralclothsimulation} processes body motion through recurrent encoders with disentangled static and dynamic branches, enabling fast cloth animation through learned temporal dynamics in latent space. However, these skeletal representations provide only body-centric motion information—they capture how the skeleton moved but not the actual cloth geometric state from previous frames. This creates ambiguity when identical poses correspond to different cloth states due to motion history. 

Test-time alignment strategies~\cite{animatablegs,ramavatar} address generalization to novel poses by mapping out-of-distribution poses back to the training distribution. Animatable-GS~\cite{animatablegs} employs PCA-based dimensionality reduction to project test poses into the training manifold, while RANA~\cite{ramavatar} introduces explicit pose space alignment through learned transformations. These methods remain effective primarily when test poses exhibit small distributional shifts from training data, as pose similarity-based alignment cannot capture motion-dependent cloth state variations.

Despite their diversity in architectural design, these learning-based methods do not establish explicit temporal causality between successive cloth states, instead learning correlations between pose patterns and appearance rather than temporal evolution rules, limiting their ability to generalize beyond training sequences.

Physics-based cloth simulation~\cite{hood,cloth3d,snug,diffcloth,position,stable,neuralclothsimulation} models garment dynamics through explicit state evolution, naturally incorporating temporal causality but bounded by manually-modeled physical constraints. Our work focuses on photorealistic avatar rendering from multi-view images, learning cloth dynamics directly from photometric observations while establishing explicit temporal causality through autoregressive geometric state propagation and distributed memory-based state retrieval.

%% file: sec/3_Method.tex
\section{Method}

\subsection{Preliminary}

\noindent\textbf{3D Gaussian Splatting.} We adopt 3D Gaussian Splatting~\cite{3dgs} as our rendering representation. Each Gaussian is parameterized by its 3D center $\mu \in \mathbb{R}^3$, rotation quaternion $q \in \mathbb{R}^4$, anisotropic scale $s \in \mathbb{R}^3$, opacity $\alpha \in \mathbb{R}$, and spherical harmonics coefficients $c$ for view-dependent color. The covariance matrix is factorized as $\Sigma = R S S^T R^T$, where $R$ is the rotation matrix derived from quaternion $q$, and $S$ is a diagonal scaling matrix. During rendering, 3D Gaussians are projected onto the 2D image plane and blended via differentiable alpha compositing.

\noindent\textbf{SMPL Model and Linear Blend Skinning.} We utilize SMPL~\cite{smpl} and SMPL-X~\cite{SMPLX} as the underlying body model, parameterized by shape $\beta \in \mathbb{R}^{10}$ and pose $\theta \in \mathbb{R}^{J \times 3}$, where $J$ denotes the number of body joints. SMPL provides a template mesh $\mathcal{T}(\beta)$ in canonical space and joint locations $\mathcal{J}(\beta)$ that enable articulated motion. To transform points from canonical space to posed space, we employ Linear Blend Skinning (LBS). Given a canonical point $x_c$ with skinning weights $w = \{w_1, \ldots, w_K\}$ and bone transformation matrices $\{B_1, \ldots, B_K\}$ derived from pose $\theta$ with $K$ joints, the posed position is:
\begin{equation}
x_p = \sum_{k=1}^K w_k B_k x_c
\end{equation}

\subsection{Overview}

We represent the avatar using a learnable Gaussian template $G_{\text{tmp}}$ in canonical space. At each frame, the network predicts deformations $\Delta G_t$ that modify the template's Gaussian attributes. These deformations are added to the template to obtain deformed canonical Gaussians, which are then transformed to posed space via Linear Blend Skinning and rendered through differentiable splatting.

Our deformation network $\Psi$ adopts a StyleUNet~\cite{StyleUnet,StyleGAN,unet} architecture and learns cloth dynamics through a dual-stream autoregressive process. As shown in \cref{fig:overview}, the geometric stream extracts multi-scale features through an encoder and aggregates them temporally using Motion-Adaptive Temporal Aggregation (MATA) to produce geometric features $A_t$. The state stream fuses $A_t$ with historical states from a memory bank to obtain temporal state $h_t$, which is decoded to predict canonical-space deformations. During training, multi-view photometric supervision and adaptive temporal regularization guide the network to discover temporal evolution patterns.

\subsection{Learnable Gaussian Template}

We model avatar dynamics by predicting deformations of a learnable Gaussian template in canonical space, which are then transformed to posed space via Linear Blend Skinning.

\noindent\textbf{Template Initialization.} We initialize an optimizable Gaussian template $G_{\text{tmp}}$ in canonical space by binding Gaussians to the SMPL-X surface using its UV parameterization. Each Gaussian is placed at a UV coordinate location with skinning weights inherited from nearby vertices. The number of Gaussians depends on the UV map resolution and remains fixed throughout training, ensuring spatial correspondence across frames.

\noindent\textbf{Template and Deformation Factorization.} We factorize the canonical Gaussians into a learnable template and frame-specific deformations:
\begin{equation}
G_t^{\text{cano}} = G_{\text{tmp}} + \Delta G_t
\end{equation}
where $\Delta G_t = \{\Delta\mu_t, \Delta q_t, \Delta s_t, \Delta\alpha_t, \Delta c_t\}$ represents deformations in Gaussian attributes. The template captures average canonical geometry and appearance, while deformations encode pose-dependent and temporally-driven variations.

\noindent\textbf{Posed Space Transformation.} To render under driving pose $\theta_t$, we transform the deformed canonical Gaussians to posed space via LBS. For a canonical Gaussian with center $\mu_c$ and covariance $\Sigma_c$:
\begin{equation}
\mu_p = \sum_{k=1}^K w_k B_k \mu_c, \quad \Sigma_p = J \Sigma_c J^T
\end{equation}
where $J$ is the rotation component of the bone transformation matrices. The posed Gaussians are rendered to the target viewpoint through splatting-based rasterization.

\subsection{Dual-Stream Autoregressive Architecture}

At each time step $t$, we predict deformations through two complementary autoregressive streams. The geometric stream explicitly conditions on the previous frame's predicted geometric deformation $\Delta\mu_{t-1}$ and processes the current temporal window through motion-adaptive aggregation. The state stream maintains a memory bank of historical temporal states and fuses them with current geometric features to capture long-term temporal evolution.

Our deformation prediction conditions on the immediate previous $\Delta\mu_{t-1}$ rather than a full historical sequence. This single-frame design is sufficient because $\Delta\mu_{t-1}$ was itself predicted from earlier frames through the autoregressive chain, thereby carrying cloth geometric history. 

\noindent\textbf{Geometric Stream.} The network input consists of UV-unwrapped position maps derived from skeletal pose and gaussian template. For each frame $\tau$ in the temporal window $\{t-n+1, \ldots, t\}$, the input comprises the position map $p_\tau$, velocity map $\Delta p_\tau$, and the geometric information $\Delta\mu_{t-1}$:
\begin{equation}
I_\tau = \text{Concat}[p_\tau, \Delta p_\tau, \Delta\mu_{t-1}]
\end{equation}
where $p_\tau$ captures surface configuration, $\Delta p_\tau = p_\tau - p_{\tau-1}$ captures motion dynamics, and $\Delta\mu_{t-1}$ is shared across all frames in the window, providing cloth geometric deformation from the previous frame. 

We extract multi-scale features through StyleUNet encoder $\Psi_{\text{enc}}$, producing an $L$-level feature pyramid for each frame, where $l=1$ denotes the finest resolution and $l=L$ the coarsest:
\begin{equation}
\{f_\tau^l\}_{\tau=t-n+1}^t, \quad l \in \{1, \ldots, L\}
\end{equation}
This yields $n$ feature pyramids, one for each frame in the temporal window. 

Motion-Adaptive Temporal Aggregation. Cloth exhibits spatially heterogeneous motion characteristics, requiring adaptive processing rather than uniform aggregation. At the deepest semantic level (Level $L$), we employ motion-adaptive temporal aggregation to process the temporal window.

We first quantify motion magnitude at each spatial location from the velocity maps to capture motion heterogeneity across different body regions:
\begin{equation}
V_\tau[u,v] = \|\Delta p_\tau[u,v]\|_2
\end{equation}
where $V_\tau[u,v]$ represents the motion magnitude at spatial location $(u,v)$ in frame $\tau$. A lightweight network $\Phi_{\text{m}}$ transforms motion magnitude maps into feature-space embeddings:
\begin{equation}
m_\tau = \Phi_{\text{m}}(V_\tau), \quad \tau \in \{t-n+1, \ldots, t\}
\end{equation}
where $\Phi_{\text{m}}$ consists of lightweight convolutions that project spatial motion patterns into the feature dimension. Following~\cite{pursuing}, we incorporate motion embeddings and temporal position encodings into the deepest features:
\begin{equation}
\tilde{f}_\tau^L = f_\tau^L + m_\tau + e_\tau, \quad \tau \in \{t-n+1, \ldots, t\}
\end{equation}
where $e_\tau$ are position embeddings. This motion-enhanced representation enables the network to attend adaptively based on local motion characteristics—high-motion regions emphasize recent dynamic information while low-motion regions maintain broader temporal context.

We then compute causal self-attention~\cite{attention,neuralattention} over the $n$ motion-enhanced frames:
\begin{equation}
Q = \Phi_Q(\{\tilde{f}_\tau^L\}), \quad K = \Phi_K(\{\tilde{f}_\tau^L\}), \quad V = \Phi_V(\{\tilde{f}_\tau^L\})
\end{equation}
where $\Phi_Q$, $\Phi_K$, $\Phi_V$ are projection networks. We apply causal masking to enforce temporal causality:
\begin{equation}
M_{\text{causal}}[\tau, \tau'] = \begin{cases}
-\infty, & \text{if } \tau < \tau' \\
0, & \text{if } \tau \geq \tau'
\end{cases}
\end{equation}
where $\tau, \tau' \in \{t-n+1, \ldots, t\}$ denote frame indices in the temporal window. The final attention output aggregates temporal information within the window while respecting causal constraints:
\begin{equation}
A_t = \text{softmax}\left(\frac{QK^T}{\sqrt{d_k}} + M_{\text{causal}}\right)V
\end{equation}
This motion guidance enables spatially-adaptive attention patterns, where actively moving regions focus on recent frames to capture transient dynamics, while stable regions distribute attention more uniformly to maintain temporal coherence. 

\noindent\textbf{State Stream.} While $A_t$ captures surface configuration and motion patterns observable from the temporal window, cloth dynamics also depend on implicit internal state that is not directly observable from geometry alone. The state stream addresses this by maintaining a memory bank of historical temporal states that encode this implicit internal state and fusing them with current geometric features. 

Specifically, we maintain a memory bank $\mathcal{M}_t = \{h_{t-m}, \ldots, h_{t-1}\}$ storing the most recent $M$ temporal states, where each $h_i$ encodes the implicit internal state from previous timesteps. At time $t$, we retrieve relevant temporal information from the memory bank and fuse it with the current geometric feature $A_t$ through cross-attention:

\begin{equation}
h^{\ast}_t = \text{CrossAttn}(Q=A_t, \, KV=\mathcal{M}_t)
\end{equation}
\begin{equation}
h_t = (1-\lambda)A_t + \lambda \cdot h^{\ast}_t
\end{equation}
where $\lambda$ is a learnable parameter that balances current geometric information and historical temporal patterns. 

After obtaining $h_t$, we update the memory bank for the next timestep:
\begin{equation}
\mathcal{M}_{t+1} = \{h_{t-M+1}, \ldots, h_{t-1}, h_t\}
\end{equation}

\noindent\textbf{Hierarchical Aggregation and Decoding.} The fused temporal state $h_t$ serves as the aggregated feature at the deepest level (Level $L$). At shallower levels (Levels 1 to $L-1$), we use lightweight convolutional networks $\Phi_{\text{agg}}^l$ for efficient temporal aggregation:
\begin{equation}
F_t^l = \begin{cases}
h_t, & l = L \\
\Phi_{\text{agg}}^l(\{f_\tau^l\}_{\tau=t-n+1}^t), & l \in \{1, \ldots, L-1\}
\end{cases}
\end{equation}
The hierarchical aggregation produces the final feature pyramid $\{F_t^l\}_{l=1}^L$, combining dual-stream fusion at the deepest semantic level with efficient convolutions at finer detail levels. The aggregated multi-scale features are fed into StyleUNet decoder $\Psi_{\text{dec}}$ with skip connections from the encoder, which predicts the canonical-space deformations:
\begin{equation}
\Delta G_t = \Psi_{\text{dec}}(\{F_t^l\}_{l=1}^L)
\end{equation}

\subsection{Adaptive Temporal Regularization}

Uniform temporal smoothness constraints create fundamental tension: they stabilize static poses but over-smooth dynamic deformations, suppressing natural momentum effects. We resolve this through motion-magnitude-aware weighting that dynamically adapts regularization strength based on spatially-varying motion characteristics.

\noindent\textbf{Motion-Magnitude-Aware Weighting.} For each Gaussian $i$ at frame $t$, we compute an adaptive weight based on its local motion magnitude. Let $(u_i, v_i)$ denote the UV coordinates of Gaussian $i$ in the canonical template. We compute per-Gaussian adaptive weights:
\begin{equation}
w_t[i] = \exp\left(-\frac{V_t[u_i, v_i]}{\tau_{\text{reg}}}\right)
\end{equation}
where $V_t[u_i, v_i]$ is the motion magnitude at the Gaussian's spatial location and $\tau_{\text{reg}}$ is a hyperparameter controlling sensitivity. This formulation achieves spatially-varying adaptive regularization: high weights for static regions enforce strong temporal consistency, while low weights for high-motion regions permit flexibility.

\noindent\textbf{Multi-Level Regularization.} We apply motion-weighted regularization at three hierarchical levels:

\textit{Feature-Level Smoothness} regularizes intermediate features at each pyramid level:
\begin{equation}
\mathcal{L}_{\text{feat}} = \sum_{l=1}^L \sum_t \gamma_l \cdot \|F_t^l - F_{t-1}^l\|_2^2
\end{equation}
where $\gamma_l=0.1$ controls regularization strength at each level. This prevents high-frequency temporal variations in learned representations that could lead to flickering artifacts.

\textit{Prediction-Level Consistency} regularizes predicted deformations with motion-adaptive weighting:
\begin{equation}
\begin{split}
\mathcal{L}_{\text{pred}} = \sum_t \sum_i w_t[i] \cdot \bigg(  \|\Delta\mu_t[i] - \Delta\mu_{t-1}[i]\|_2^2  &+ \|\Delta q_t[i] - \Delta q_{t-1}[i]\|_2^2 \\
& + \|\Delta s_t[i] - \Delta s_{t-1}[i]\|_2^2 \bigg)
\end{split}
\end{equation}
The spatially-varying weights $w_t[i]$ enforce strong consistency in static regions (e.g., torso during standing) while permitting flexibility in dynamic regions (e.g., sleeves during arm swinging).

\textit{Cross-Frame Position Smoothness} penalizes second-order temporal differences in posed positions:
\begin{align}
\text{accel}_t[i] &= \mu_{t+1}^{\text{p}}[i] - 2\mu_t^{\text{p}}[i] + \mu_{t-1}^{\text{p}}[i] \\
\mathcal{L}_{\text{sm}} &= \sum_t \sum_i w_t[i] \cdot \|\text{accel}_t[i]\|_2^2
\end{align}
where $\mu_t^{\text{p}}[i]$ denotes the posed position of Gaussian $i$ at frame $t$. This acceleration penalty minimizes jerk for smooth trajectories while motion-adaptive weighting permits natural momentum-driven motion in active regions.

\noindent\textbf{Training Objective.} Our complete training loss combines multi-view photometric supervision with adaptive temporal regularization:
\begin{equation}
\mathcal{L}_{\text{total}} = \mathcal{L}_{\text{render}} + \lambda_{\text{feat}} \mathcal{L}_{\text{feat}} + \lambda_{\text{pred}} \mathcal{L}_{\text{pred}} + \lambda_{\text{sm}} \mathcal{L}_{\text{sm}}
\end{equation}
The rendering loss comprises:
\begin{equation}
\mathcal{L}_{\text{render}} = \lambda_{\text{rgb}}\mathcal{L}_{\text{rgb}}+\lambda_{\text{ssim}}\mathcal{L}_{\text{ssim}}+\lambda_{\text{lpips}}\mathcal{L}_{\text{lpips}}
\end{equation}
where $\mathcal{L}_{\text{rgb}}$ is the L1 loss between rendered and ground-truth images, $\mathcal{L}_{\text{ssim}}$~\cite{SSIM} measures structural similarity, and $\mathcal{L}_{\text{lpips}}$~\cite{LPIPS} captures perceptual quality. These photometric losses provide supervision from multiple camera views, enabling the network to learn accurate cloth deformations from visual observations alone without requiring explicit geometric supervision.

%% file: sec/4_Experiment.tex
\section{Experiments}
\label{sec:experiments}

\subsection{Experimental Setup}

\noindent\textbf{Datasets.} We evaluate our method on 4D-DRESS~\cite{4ddress} and AvatarREX~\cite{avatarrex}. Each subject contains 5-6 motion sequences (approximately 1000 frames total, captured by 8 calibrated cameras). We train on 3-4 sequences and test on a separate held-out sequence. To further validate generalization to challenging out-of-distribution poses, we additionally evaluate on motion sequences from AMASS~\cite{amass}, which provides diverse motion patterns including sports, dancing, and acrobatic movements substantially different from training data.

\noindent\textbf{Distribution Divergence Quantification.} To systematically evaluate generalization under varying distribution shifts, we quantify pose distribution divergence between training and test sets using Maximum Mean Discrepancy (MMD)~\cite{mmd}. Given pose feature sets $\mathcal{X}_{\text{train}}$ and $\mathcal{X}_{\text{test}}$ extracted from SMPL-X parameters, MMD measures distributional distance via kernel embeddings:
\begin{equation}
\begin{aligned}
\text{MMD}^2 = &\|\mu_{\mathcal{X}} - \mu_{\mathcal{Y}}\|^2_{\mathcal{H}}\\
= &\mathbb{E}[k(x,x')] - 2\mathbb{E}[k(x,y)] + \mathbb{E}[k(y,y')]
\end{aligned}
\end{equation}
where $k$ is an RBF kernel and $\mathcal{H}$ is the reproducing kernel Hilbert space. We categorize test sequences as Near-Distribution (ND) with MMD $< 0.3$ (similar motion patterns) or Far-Distribution (FD) with MMD $> 0.6$ (substantially different dynamics), enabling rigorous evaluation of out-of-distribution generalization.

\noindent\textbf{Baselines.} We compare against HumanNeRF~\cite{humannerf}, GaussianAvatar~\cite{gaussianavatar}, and Animatable-GS~\cite{animatablegs}, representing NeRF and Gaussian Splatting paradigms. All baselines use official implementations trained on identical data with recommended settings.

\noindent\textbf{Metrics.} We evaluate rendering quality using PSNR, SSIM~\cite{SSIM} and LPIPS~\cite{LPIPS}.

\noindent\textbf{Implementation Details.} We employ a StyleUNet architecture with temporal window size $n=10$, feature pyramid levels $L=4$, UV map resolution $H\times W=512\times512$, $\sim$200k Gaussians per subject, memory bank size $m=10$ frames. Training uses Adam optimizer (learning rate $lr=6 \times 10^{-4}$). Loss weights: $\lambda_{\text{render}} = 1$, $\lambda_{\text{ssim}} = 0.2$, $\lambda_{\text{lpips}} = 0.5$, $\lambda_{\text{feat}} = 0.01$, $\lambda_{\text{pred}} = 0.1$, $\lambda_{\text{sm}} = 0.05$.

\subsection{Comparisons}

Table~\ref{tab:quantitative} presents quantitative evaluation. Our method consistently outperforms baselines, confirming that explicit modeling of kinematic information and implicit internal state enables robust generalization.

\begin{figure*}[t]
\centering
\includegraphics[width=\textwidth]{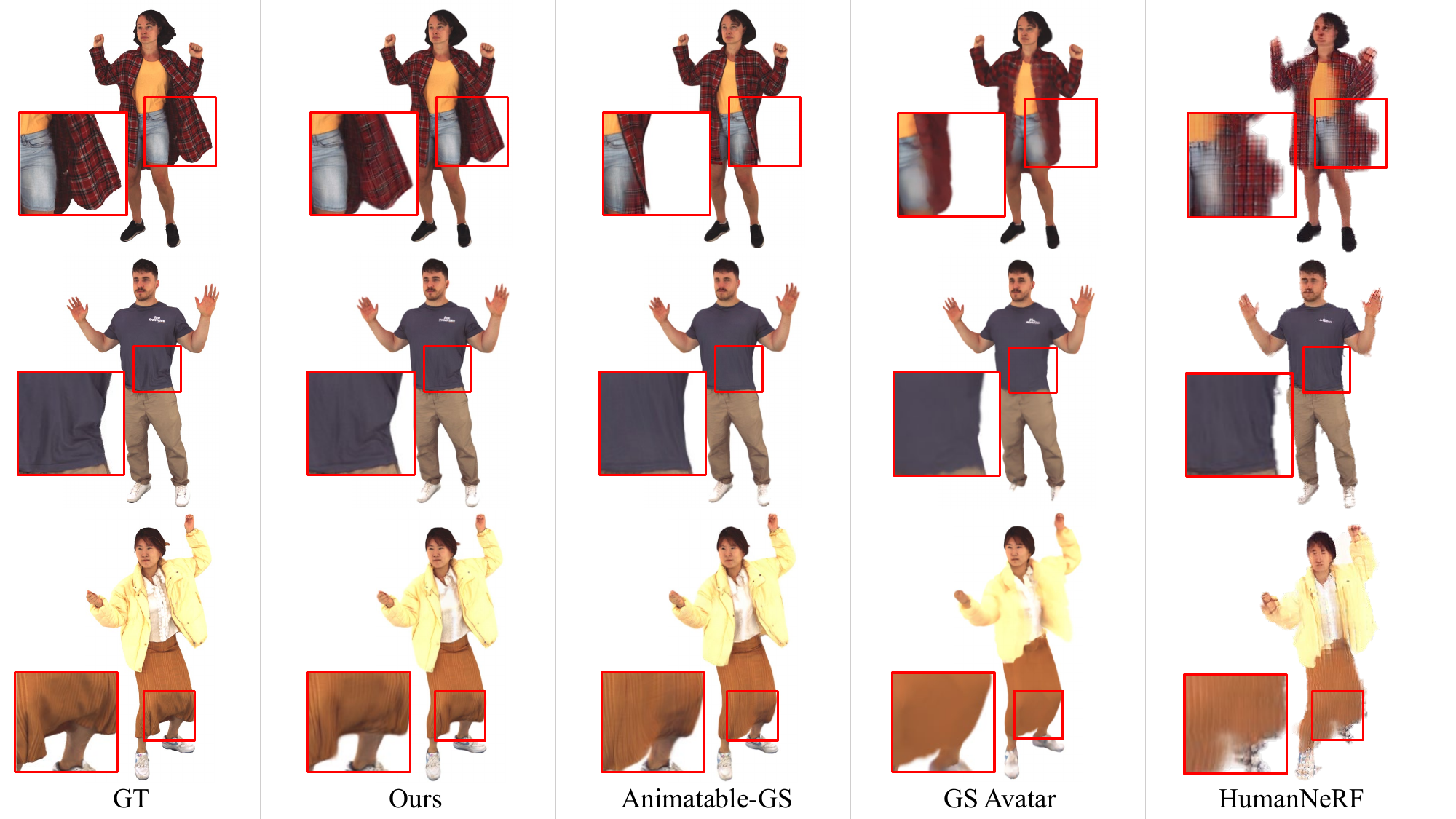}
\caption{Qualitative comparison with other methods on challenging test poses. Our method better captures momentum-driven cloth dynamics and generalizes to novel poses.}
\label{fig:comparison}
\end{figure*}

Figure~\ref{fig:comparison} demonstrates our method's ability to capture realistic cloth dynamics. In post-rotation stopping (row 1), our method correctly models momentum-driven evolution: the jacket continues swinging after the body stops, then gradually settles. Baselines produce artifacts due to instantaneous modeling—without explicit kinematic feedback and internal state encoding, they infer appearance solely from current pose. Similar improvements appear in arm movements (row 2) and jumping (row 3), where our dual-stream architecture captures natural dynamics while baselines exhibit flickering and over-smoothing.

\subsection{Ablation Studies}
\begin{figure}[t]
\centering
\includegraphics[width=0.8\columnwidth]{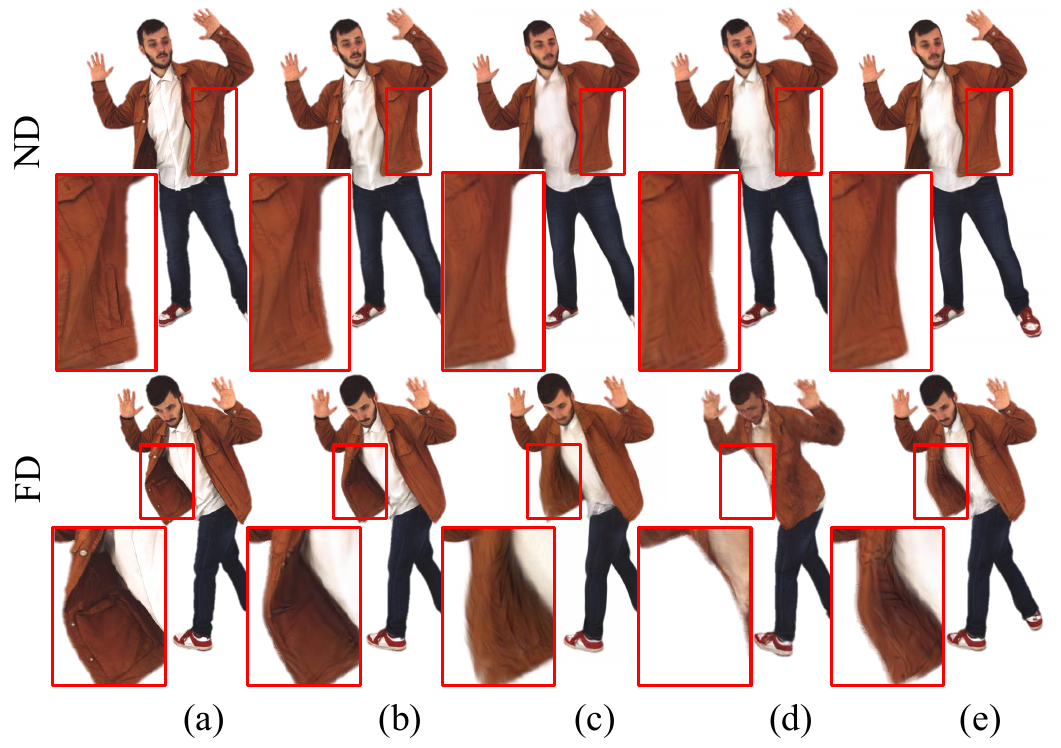}
\caption{Ablation study. (a) Ground Truth, (b) Full model with both streams, (c) w/o adaptive regularization, (d) w/o Geometric Stream, (e) w/o State Stream.}
\label{fig:ablation}
\end{figure}

We systematically validate each component's contribution through ablation studies. Table~\ref{tab:ablation} and Figure~\ref{fig:ablation} compare five variants: (a) Ground Truth, (b) Full model, (c) w/o Adaptive Regularization, (d) w/o Geometric Stream (removing $\Delta\mu_{t-1}$ input), (e) w/o State Stream (removing memory bank and cross-attention).

Results reveal the complementary necessity of both streams, particularly on far-distribution (FD) poses. The w/o Geometric Stream variant (d) exhibits the most severe degradation, confirming that explicit geometric deformation feedback is essential—without $\Delta\mu_{t-1}$ and motion-adaptive aggregation, the network must infer cloth configuration and motion from body pose alone, leading to accumulated errors in position and velocity estimation. The w/o State Stream variant (e) also degrades significantly, demonstrating that implicit internal state cannot be reliably recovered from body pose alone.

Notably, the substantially larger performance degradation on FD versus ND sets validates that explicit temporal causality modeling is essential for achieving robust generalization beyond the training distribution.

\begin{table}[t]
\centering
\caption{Quantitative comparison. All methods are evaluated at 940$\times$1280 resolution.}
\label{tab:quantitative}
\small
\begin{tabular*}{\columnwidth}{@{\extracolsep{\fill}}l|cccc@{}}
\toprule
Method & PSNR$\uparrow$ & SSIM$\uparrow$ & LPIPS$\downarrow$\\
\midrule
HumanNeRF & 20.0567 & 0.9121 & 0.1292 \\
GaussianAvatar & 26.2311 & 0.9575 & 0.0715 \\
Animatable-GS & \underline{29.5138} & \underline{0.9724} & \underline{0.0413} \\
\midrule
Ours & \textbf{31.0765} & \textbf{0.9839} & \textbf{0.0305} \\
\bottomrule
\end{tabular*}
\vspace{-0.1in}
\end{table}

\begin{table}[t]
\centering
\caption{Ablation study on ND (MMD=0.23) and FD (MMD=0.87) test sequences.}

\label{tab:ablation}
\small
\begin{tabular*}{\columnwidth}{@{\extracolsep{\fill}}l|l|ccc@{}}
\toprule
Split & Variant & PSNR$\uparrow$ & SSIM$\uparrow$ & LPIPS$\downarrow$ \\
\midrule
\multirow{4}{*}{ND} & Full Model & \textbf{31.7153} & \textbf{0.9802} & \textbf{0.0283} \\
& w/o Adaptive Reg. & 29.3130 & \underline{0.9680} & 0.0453 \\
& w/o GeoStream. & 28.9622 & 0.9651 & 0.0476 \\
& w/o StateStream. & \underline{30.2593} & 0.9676 & \underline{0.0386} \\
\midrule
\multirow{4}{*}{FD} & Full Model & \textbf{30.6720} & \textbf{0.9739} & \textbf{0.0302} \\
& w/o Adaptive Reg. & \underline{28.8778} & \underline{0.9658} & \underline{0.0492} \\
& w/o GeoStream. & 24.2694 & 0.9557 & 0.0623 \\
& w/o StateStream. & 28.2286 & 0.9632 & 0.0511 \\
\bottomrule
\end{tabular*}
\vspace{-0.1in}
\end{table}

%% file: sec/5_Conclusion.tex
\section{Conclusion}

We present a dual-stream autoregressive framework for temporally coherent animatable avatars. We identify that cloth rendering requires modeling both observable kinematic information and implicit internal state—dual requirements that skeletal poses alone cannot satisfy. By explicitly modeling kinematic information through autoregressive conditioning and implicit internal state through memory-based fusion, our method learns temporally coherent, photorealistic cloth dynamics directly from multi-view photometric observations.

Experiments demonstrate robust generalization to motion patterns beyond training distributions, validating that dual-stream temporal modeling is essential for realistic cloth dynamics in neural rendering.

%% file: Supplementary_Material.tex
\section{Supplementary Material}

In this supplementary material, we provide additional implementation details 
and experimental results to complement the main paper. We begin by presenting 
the detailed network architecture in Sec.~\ref{sec:supp_arch}. Following that, 
we elaborate on our two-stage training strategy in Sec.~\ref{sec:supp_training}, 
showcase more avatar results in Sec.~\ref{sec:supp_results}, and present 
additional experiments including ablation studies and efficiency comparisons 
in Sec.~\ref{sec:supp_additional}. We further provide fine-grained component 
ablations and temporal consistency evaluations in Sec.~\ref{sec:supp_more_experiments}. 
Finally, we discuss the limitations of our approach and outline directions 
for future work in Sec.~\ref{sec:supp_limitations}.

\subsection{Network Architecture}
\label{sec:supp_arch}

\textbf{StyleUNet Architecture.} Our deformation network $\Psi$ adopts a StyleUNet~\cite{StyleUnet} architecture with input channel dimension $C_{in} = 9$ (position map $p_\tau$: 3 channels, velocity map $\Delta p_\tau$: 3 channels, previous deformation $\Delta\mu_{t-1}$: 3 channels). The encoder $\Psi_{\text{enc}}$ consists of four downsampling blocks, producing a 4-level feature pyramid with channels $\{64, 128, 256, 512\}$ at resolutions $\{512 \times 512, 256 \times 256, 128 \times 128, 64 \times 64\}$. Each frame $\tau$ in the temporal window is processed independently through the encoder. The decoder $\Psi_{\text{dec}}$ employs upsampling blocks with skip connections from the encoder to predict Gaussian deformations $\Delta G_t = \{\Delta\mu_t, \Delta q_t, \Delta s_t, \Delta\alpha_t, \Delta c_t\}$.

\textbf{Motion-Adaptive Temporal Aggregation (MATA).} The motion embedding network $\Phi_m$ employs convolutional layers to project motion magnitude maps $V_\tau$ into feature space matching Level 4 dimension (512 channels). At the deepest level ($L=4$), causal self-attention with 8 heads aggregates motion-enhanced features $\{\tilde{f}_\tau^L\}_{\tau=t-n+1}^t$ to produce geometric features. At shallower levels ($l \in \{1, 2, 3\}$), lightweight temporal convolutions provide efficient aggregation.

\textbf{Memory Bank and State Stream.} The memory bank $\mathcal{M}_t = \{h_{t-m}, \ldots, h_{t-1}\}$ stores $m=10$ most recent temporal states at Level 4 resolution ($64 \times 64 \times 512$). Cross-attention fuses geometric features from MATA with historical states from the memory bank, using 8 attention heads.

\subsection{Training Strategy}
\label{sec:supp_training}

We employ a two-stage optimization strategy. \textbf{Stage 1:} Warm-up phase without autoregressive connections by setting all $\Delta\mu_{t-1} = 0$ and initializing memory bank $\mathcal{M}_t = h_1$ with loss $\mathcal{L}_{\text{stage}_1} = \mathcal{L}_{\text{render}}$. This establishes stable feature extraction before introducing temporal dependencies. \textbf{Stage 2:} Full autoregressive training with geometric stream (using predicted $\Delta\mu_{t-1}$) and state stream (updating memory bank with $h_t$) enabled: $\mathcal{L}_{\text{stage}_2} = \mathcal{L}_{\text{render}} + \lambda_{\text{feat}}\mathcal{L}_{\text{feat}} + \lambda_{\text{pred}}\mathcal{L}_{\text{pred}} + \lambda_{\text{sm}}\mathcal{L}_{\text{sm}}$. We use Adam optimizer with learning rate $6 \times 10^{-4}$. Loss weights: $\lambda_{\text{render}} = 1$, $\lambda_{\text{ssim}} = 0.2$, $\lambda_{\text{lpips}} = 0.5$, $\lambda_{\text{feat}} = 0.01$, $\lambda_{\text{pred}} = 0.1$, $\lambda_{\text{sm}} = 0.05$.

\subsection{Additional Avatar Results}
\label{sec:supp_results}

\begin{figure*}[htbp]
\centering
\includegraphics[width=\linewidth]{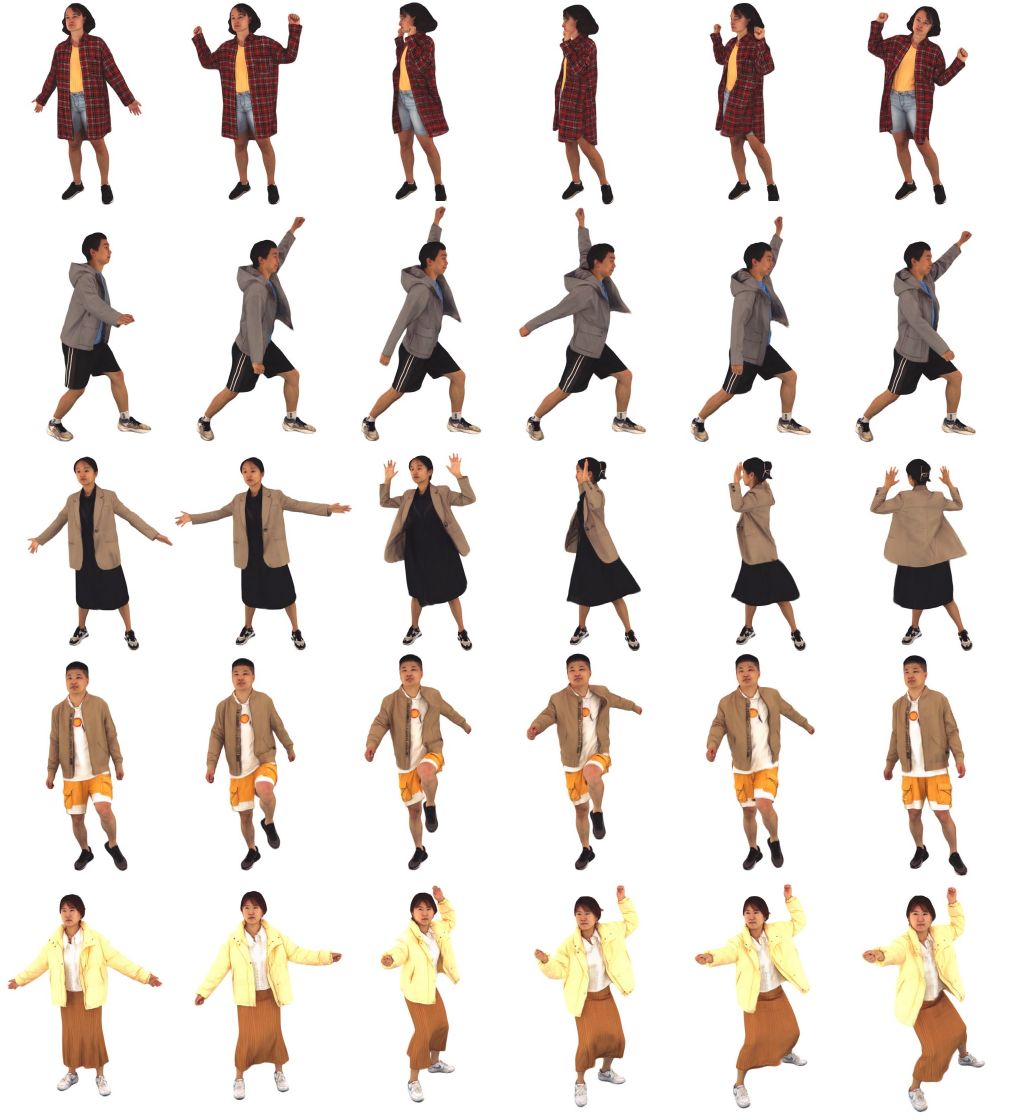}
\caption{\textbf{Avatar gallery.} Additional results demonstrating our method's capability to reconstruct diverse avatars with different clothing styles and motion patterns. Our dual-stream architecture captures realistic cloth dynamics including momentum-driven motion and natural settling behaviors.}
\label{fig:supp_gallery}
\end{figure*}

Figure~\ref{fig:supp_gallery} showcases additional avatar reconstruction results across different subjects and motion sequences from our dataset. Our method successfully captures diverse clothing styles including loose jackets, flowing dresses, and multi-layer garments, demonstrating robust generalization across various scenarios. The results exhibit temporally coherent cloth dynamics with natural momentum effects, validating the effectiveness of our dual-stream autoregressive approach.

\subsection{Additional Experiments}
\label{sec:supp_additional}

\textbf{Ablation on Temporal Window and Memory Bank Size.}
We investigate the joint impact of temporal window size $n$ (in the geometric stream) and memory bank size $m$ (in the state stream) on reconstruction quality and training efficiency. We evaluate three configurations: $(n=5, m=5)$, $(n=10, m=10)$ (our default), and $(n=20, m=20)$ on subject ``4D-DRESS\_00175\_Outer\_2". Table~\ref{tab:supp_window} presents the quantitative comparison.

\begin{table}[h]
\centering
\caption{\textbf{Ablation on temporal window and memory bank size.} Comparison of different configurations where both temporal window size $n$ and memory bank size $m$ are varied jointly. Our default setting uses $n=10, m=10$.}
\label{tab:supp_window}
\small
\begin{tabular}{lcccc}
\toprule
$(n, m)$ & PSNR$\uparrow$ & SSIM$\uparrow$ & LPIPS$\downarrow$ & Training (h)$\downarrow$ \\
\midrule
$(5, 5)$ & 30.2421 & 0.9791 & 0.0331 & \textbf{20}\\
$(10, 10)$ & \textbf{30.9515} & \underline{0.9798} & \textbf{0.0311} & \underline{28} \\
$(20, 20)$ & \underline{30.8193} & \textbf{0.9801} & \underline{0.0314} & 35 \\
\bottomrule
\end{tabular}
\end{table}

The results show that $(n=10, m=10)$ achieves the best balance across rendering quality and computational efficiency. While $(n=10, m=10)$ and $(n=20, m=20)$ demonstrate comparable rendering performance, the larger configuration substantially increases training time without meaningful quality improvements. The smaller configuration $(n=5, m=5)$ reduces training time but slightly degrades rendering quality due to insufficient temporal context---both the geometric stream lacks sufficient motion history to capture dynamics patterns, and the state stream has limited historical states for retrieving relevant temporal information. The balanced configuration $(n=10, m=10)$ provides sufficient historical information for both streams to effectively model cloth dynamics, validating our choice as the optimal trade-off between temporal modeling capacity and computational cost.

\textbf{Rendering Efficiency Comparison.}
We compare the rendering efficiency of our method against baseline approaches. Table~\ref{tab:supp_fps} reports frames per second (FPS) measurements at $940 \times 1280$ resolution on an NVIDIA RTX 4090 GPU. All methods use their official implementations with recommended settings.

\begin{table}[h]
\centering
\caption{\textbf{Rendering efficiency comparison.} FPS at $940 \times 1280$ resolution on RTX 4090.}
\label{tab:supp_fps}
\small
\begin{tabular*}{\columnwidth}{@{\extracolsep{\fill}}lcccc@{}}
\toprule
 & HumanNeRF~\cite{humannerf} & GaussianAvatar~\cite{gaussianavatar} & Animatable-GS~\cite{animatablegs} & Ours \\
\midrule
FPS & 0.18 & \textbf{35} & \underline{13} & 10 \\
\bottomrule
\end{tabular*}
\end{table}

Our method achieves fast rendering performance while providing superior rendering quality and temporal coherence through dual-stream autoregressive modeling. The additional computational cost from temporal window processing and memory bank retrieval is modest compared to the significant improvement in temporal consistency.

\begin{figure}[h]
\centering
\includegraphics[width=0.7\linewidth]{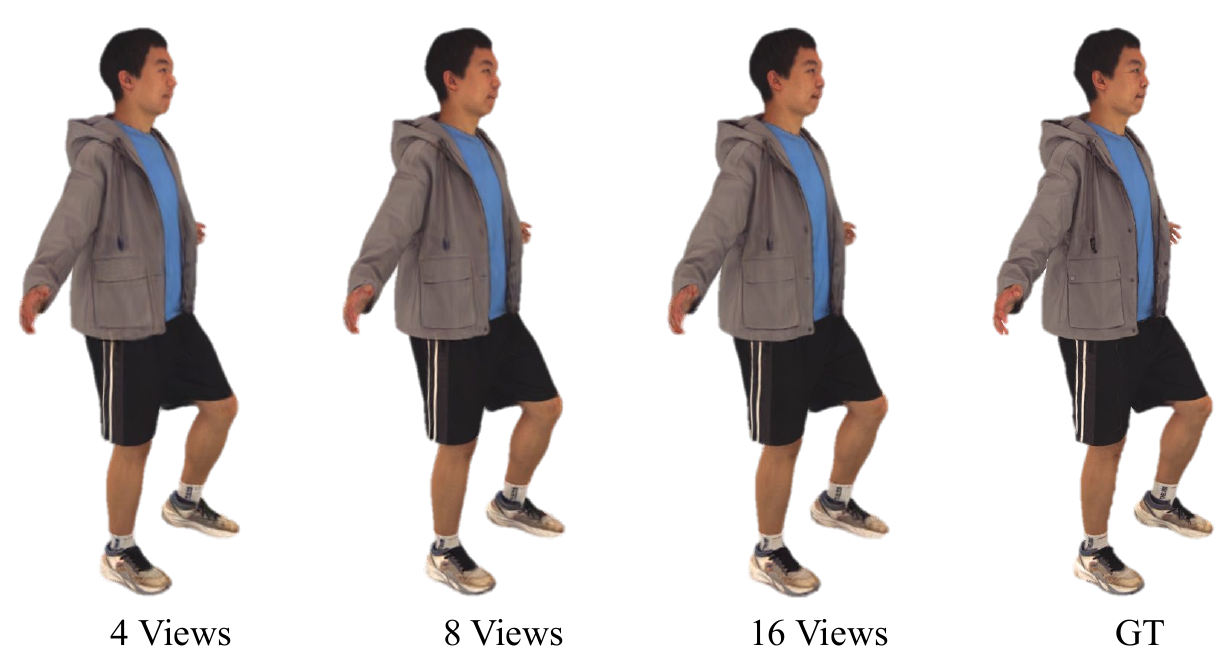}
\caption{\textbf{Impact of camera views.} Visual comparison of reconstruction quality with different numbers of training views.}
\label{fig:supp_views}
\end{figure}

\textbf{Impact of Camera Views.}
We investigate the robustness of our method under varying numbers of training views. To systematically evaluate the impact of camera view count, we artificially render different numbers of camera viewpoints (4, 8, and 16 views) using the ground truth texture data provided by 4D-DRESS. We evaluate on subject ``4D-DRESS\_00135\_Outer\_2". Table~\ref{tab:supp_views} presents quantitative results, and Figure~\ref{fig:supp_views} shows qualitative comparisons.

\begin{table}[h]
\centering
\caption{\textbf{Impact of camera views.} Quantitative evaluation with different numbers of training views.}
\label{tab:supp_views}
\small
\begin{tabular}{lccc}
\toprule
Views & PSNR$\uparrow$ & SSIM$\uparrow$ & LPIPS$\downarrow$ \\
\midrule
4 & 30.5542 & 0.9791 & 0.0318 \\
8 & \underline{31.0515} & \underline{0.9805} & \underline{0.0313} \\
16 & \textbf{31.2893} & \textbf{0.9817} & \textbf{0.0296} \\
\bottomrule
\end{tabular}
\end{table}

The results demonstrate that our method achieves robust and consistent performance across different camera configurations. Our dual-stream temporal modeling effectively captures cloth dynamics even with limited observational constraints, as the explicit temporal causality (through autoregressive geometric propagation and memory-based state retrieval) provides strong regularization that compensates for sparse multi-view supervision. This makes the method practical for real-world capture scenarios with sparse camera arrays.

\subsection{More Experiments}
\label{sec:supp_more_experiments}

\textbf{Quantitative Comparison with Temporal Consistency.}
In addition to PSNR and SSIM, we report Fr\'{e}chet Video Distance (FVD) to evaluate temporal consistency. Table~\ref{tab:supp_temporal_main} compares our method with HumanNeRF, GaussianAvatar, Animatable-GS, MMLPs, and D3GA. Our method achieves the best rendering quality and the lowest FVD, indicating substantially improved temporal coherence.

\begin{table}[h]
\centering
\caption{\textbf{Quantitative comparison with FVD.} Lower FVD indicates better temporal consistency.}
\label{tab:supp_temporal_main}
\small
\begin{tabular}{lccc}
\toprule
Method & PSNR$\uparrow$ & SSIM$\uparrow$ & FVD$\downarrow$ \\
\midrule
HumanNeRF & 20.0567 & 0.9121 & 20.24 \\
GaussianAvatar & 26.2311 & 0.9575 & 12.34 \\
Animatable-GS & 29.5138 & 0.9724 & 6.17 \\
MMLPs (CVPR 2025) & 29.6843 & 0.9731 & 4.93 \\
D3GA (3DV 2025) & 29.7215 & 0.9728 & 3.38 \\
\textbf{Ours} & \textbf{31.0765} & \textbf{0.9839} & \textbf{0.31} \\
\bottomrule
\end{tabular}
\end{table}

\textbf{Fine-Grained Component Ablation.}
We further evaluate individual design choices on the FD split. Table~\ref{tab:supp_finegrained} shows that removing previous deformation feedback, velocity maps, causal attention, motion embeddings, or the state stream degrades both image quality and temporal consistency. Replacing the state stream with a long skeletal pose history remains substantially inferior, confirming that pose history cannot substitute for geometric feature memory. Extending the deformation-conditioning window beyond the immediate previous deformation yields negligible gain, supporting the single-step autoregressive design because $\Delta\mu_{t-1}$ already carries history through the autoregressive chain.

\begin{table}[h]
\centering
\caption{\textbf{Fine-grained component ablation on the FD split.}}
\label{tab:supp_finegrained}
\small
\begin{tabular}{lccc}
\toprule
Variant & PSNR$\uparrow$ & SSIM$\uparrow$ & FVD$\downarrow$ \\
\midrule
\textbf{Full Model} & \textbf{30.6723} & \textbf{0.9739} & \textbf{0.23} \\
w/o $\Delta\mu_{t-1}$ & 24.2694 & 0.9557 & 12.76 \\
w/o velocity map $\Delta p$ & 29.8436 & 0.9682 & 1.87 \\
w/o both (position map only) & 23.8318 & 0.9498 & 13.83 \\
Bidirectional attn (no causal) & 28.3341 & 0.9648 & 5.31 \\
Direct cross-attn on prev. frame & 29.1247 & 0.9661 & 2.14 \\
w/o motion embedding & 30.0153 & 0.9703 & 1.94 \\
w/ long pose hist. (w/o State Stream) & 28.3174 & 0.9637 & 6.86 \\
\midrule
$\Delta\mu_{t-1}$ (ours) & 30.6720 & 0.9739 & \textbf{0.23} \\
$\Delta\mu_{t-1:t-3}$ & \textbf{30.7161} & \textbf{0.9741} & 0.25 \\
$\Delta\mu_{t-1:t-5}$ & 30.5247 & 0.9733 & 0.29 \\
\bottomrule
\end{tabular}
\end{table}

\subsection{Limitations and Future Work}
\label{sec:supp_limitations}

\textbf{Limitations.} Our approach operates in a per-subject manner, requiring separate training for each clothed human, which limits scalability when dealing with diverse individuals and garment types. Additionally, the UV parameterization assumes continuous surface topology and cannot naturally handle clothing tearing, cutting, or topological changes. The autoregressive nature may also accumulate small errors over very long sequences, though our adaptive regularization strategy provides stabilization.

\textbf{Future Work.} Several directions could extend our framework. Exploring generalizable models that learn shared priors across subjects could eliminate per-subject training while preserving personalized detail. Furthermore, extending the representation to handle topological changes would enable modeling of garment damage and complex interactions. The dual-stream principle is also conceptually transferable to other deformation-based dynamic representations, which we leave as future work.